\documentclass[letterpaper, 10 pt, conference]{ieeeconf}  % Comment this beam out if you need a4paper

\IEEEoverridecommandlockouts                              % This command is only needed if 
\usepackage{amsmath}
\usepackage{graphicx}
\usepackage{color}
\newcounter{RomanNumber}

\newcommand{\Reffig}[1]{Fig.~\ref{#1}}
\newcommand{\Refsec}[1]{Sec.~\ref{#1}}
\newcommand{\Reftab}[1]{Tab.~\ref{#1}}

\title{\LARGE \bf
DLO: Direct LiDAR Odometry for 2.5D Outdoor Environment
}

\author{Lu Sun, Junqiao Zhao$^{*}$, Xudong He, Chen Ye% <-this % stops a space
\thanks{This work is supported by the National Natural Science Foundation of China (No. U1764261), the Natural Science Foundation of Shanghai (No.kz170020173571) and the Fundamental Research Funds for the Central Universities (No. 22120170232, No. 22120180095).}% <-this % stops a space
\thanks{All authors are with the Department of Computer Science and Technology, School of Electronics and Information Engineering, Tongji University, Shanghai, 201804, China
        {\tt\small zhaojunqiao@tongji.edu.cn}}%
}

\begin{document}

\maketitle
\thispagestyle{empty}
\pagestyle{empty}

%%%%%%%%%%%%%%%%%%%%%%%%%%%%%%%%%%%%%%%%%%%%%%%%%%%%%%%%%%%%%%%%%%%%%%%%%%%%%%%%
\begin{abstract}
For autonomous vehicles, high-precision real-time localization is the guarantee of stable driving.
Compared with the visual odometry (VO), the LiDAR odometry (LO) has the advantages of higher accuracy and better stability. 
However, 2D LO is only suitable for the indoor environment, and 3D LO has less efficiency in general.
Both are not suitable for the online localization of an autonomous vehicle in an outdoor driving environment. 
In this paper, a direct LO method based on the 2.5D grid map is proposed. 
The fast semi-dense direct method proposed for VO is employed to register two 2.5D maps. 
Experiments show that this method is superior to both the 3D-NDT and LOAM in the outdoor environment.

\end{abstract}

%%%%%%%%%%%%%%%%%%%%%%%%%%%%%%%%%%%%%%%%%%%%%%%%%%%%%%%%%%%%%%%%%%%%%%%%%%%%%%%%
\section{INTRODUCTION}
 % For autonomous vehicles, high-precision localization is the guarantee of stable driving. 
 In the absence of GNSS, an autonomous vehicle must have the ability to locate itself through the sensors' perception of the environment. 
 The Simultaneous Localization and Mapping (SLAM) algorithm constructs a map by observing environmental landmarks during driving, and correlates and matches the current perception to the map in real time to optimally estimate the ego-pose.

 In a SLAM system, one of the most important parts is the front-end, which usually performs the frame-by-frame pose estimation, resulting in an odometry sub-system.
 % The front-end determines the stability and accuracy of the entire system. 
 Visual odometry (VO) and LiDAR odometry (LO) both have been intensively explored.
 % the most commonly used front-end. 
 The VO methods can be classified into indirect and direct method \cite{engel2017direct}.
 % using the camera has characteristics such as low cost, low data amount, easy extraction features and the like. 
 The indirect methods rely on features extracted from images, such as the ORB feature adopted in ORB-SLAM \cite{mur2015orb}.
 The transformation between consequent keyframes is then calculated by the matching of features.
 The direct methods, such as LSD-SLAM \cite{engel2014lsd}, directly estimate transformation between keyframes based on the minimization of the photometric error between two images, thus are faster and more robust than indirect methods in the "texture-less" scene which contains little features.
 The combination of both merits is also proposed by the SVO \cite{forster2014svo}.
However, as the camera is easily affected by the changing illumination, VO methods still suffer from robust issues, especially in the outdoor environment.

Thanks to the active sensing and the high precision of LiDAR sensor, the LO methods are more robust and accurate than VO in general.
The existing LO methods can be classified based on the spatial dimensionality of the measurement involved. 
2D LO methods use a two-dimensional occupancy grid map (OGM) with branch-and-bound \cite{Lawler1966Branch} or gradient descending scan-match method to estimate the sensor's motion. 
Practical systems based on these methods are GMapping \cite{grisetti2007improved}, Hector SLAM \cite{KohlbrecherMeyerStrykKlingaufFlexibleSlamSystem2011}, and cartographer \cite{hess2016real}. 
% Because of the low data amount of the single-beam LiDAR, 2D LO can have a very high frequency. 
However, due to the insufficient representative power of the 2D OGM for outdoor environments, these methods are most suitable for the indoor environment.

The 3D LO methods employ point clouds from the multi-beam LiDAR to estimate the motion of the sensor by matching of point clouds \cite{besl1992method} or 3D OGMs \cite{magnusson2009three}, or by matching of 3D features extracted from point clouds \cite{zhang2014loam}.
 % 3D information of the environment. 
Because of the accessible 3D information, the 3D LO method can be applied to both indoor and outdoor environments.
 % contains the height information of the obstacle points, which can be accurately estimated in outdoor environments and can be applied to scenes with height changes, such as uphill and downhill. 3D LO has two kinds of implement: ICP \cite{besl1992method}, LOAM \cite{zhang2014loam} based on raw data iteration and 3D-NDT based on 3D voxel grid. Both of these methods can accurately calculate the transition matrix of two adjacent frames of the point cloud, but due to the huge amount of data, the system is difficult to be real-time.
Nevertheless, the expensive computation required for point cloud registration or OGM matching in 3D space makes these methods hardly fit the online localization.

The key insight of this paper is that the height variation in the surrounding of a vehicle (represented by a 2.5D grid map or a height map) is discriminative and lightweight, compared to the above two popular representations.
Although 2.5D-based localization methods already exist  \cite{wolcott2015fast}, they cannot be used as an odometry system since a pre-built map is mandatory.

%

% and in most of the time autonomous vehicles travel on planes without height changes. Therefore, we hope to use the 3D information to help autonomous vehicles' two-dimensional localization in the outdoor environment. This will not only ensure the real-time system but also ensure the use of enough environmental information.
This paper presents an LO method based on the 2.5D grid map for autonomous vehicles in the outdoor environment. 
The main contribution is the usage of the semi-dense direct method originated from VO to realize a fast and accurate registration of 2.5D representations of LiDAR scans.

Firstly, a 2.5D grid map with each cell retaining a height expectation is built. 
Then, cells with high gradient are selected, and Gauss-Newton method is used to optimize the objective function based on the height difference error (HDE) of two 2.5D grid maps.
Finally, the optimal transition matrix of two maps is obtained. 
The experimental results showed the proposed method can register two frames of HDL-64 LiDAR in centimeter accuracy in 40 ms, comparing to 39s of 3D-NDT and 0.93s of LOAM (without IMU).
 % % point cloud in a 2D plane. 
 % With this method, a fast and stable 2D LO can be realized. By experimenting on the KITTI dataset and the actual autonomous vehicle platform, it can be proved that this method has short running time and high precision, and fully meets the front-end requirement of an autonomous vehicle SLAM system.

\section{RELATED WORK}

  % In recent years, simultaneous localization and mapping (SLAM) algorithm have drawn more and more researchers' attention. SLAM methods are usually divided into two functional parts: front-end and back-end. The front end mainly observes the environment by a sensor, sequentially predict the location of the sensor and its observed features or landmarks. At present, the SLAM front-end mainly implement based on LiDAR and vision. Among them, SLAM based on LiDAR can continuously and proactively explore the environment, obtain the discrete sampling (3D point cloud) in 3D coordinates of the corresponding scene, and some can also obtain the surface reflection intensity of objects in the scene. Due to the 3D point cloud usually has larger density and higher positional accuracy, we can regard it as the observation result, and estimate the sensor motion(rotation and translation) between two observations using a point cloud registration algorithm.
Existing LO or LiDAR SLAM methods can be divided into 2D, 3D, and 2.5D, according to their map representation. 
Since the 2D-based methods could not be applied in many open outdoor environments, where merely 2D occupancy information is not sufficient, this section will focus on the 3D and the 2.5D based methods.
% The most commonly used method is the 2D grid map \cite{elfes1989using}. The 2d grid map refers to project the 3d lidar point cloud into the 2d plane by applying its corresponding x,y position relation. At present, the best 2d lo is cartographer \cite{hess2016real} made by Google in 2016. It achieved very excellent performance in indoor environments, but it hardly achieves the same results in outdoor due to its limited 2d lidar information.

The classic ICP algorithm \cite{besl1992method}, initially proposed by Besl and Mckey, is a point set-based registration method.
The registration process is to iteratively find the optimal transition matrix between two point clouds based on the point-wise distance metric. 
However, multiple iterations for searching the nearest point pair resulting in a high computational cost. And the result of ICP strongly dependent on the initial value.

%

%To take into account of the full 3D information, the 3D voxel map is proposed \cite{kaufman1993volume}.
 % which contains rich scene information such as height and reflectivity. 
The 3D-NDT \cite{magnusson2009three} algorithm partitions the point cloud into 3D voxel-based representation.
The multi-variable normal distribution in x, y, and z-axis for points in each cell is calculated.
The probability sum of the target point cloud in the distributions of the referencing voxel-based representation subject to certain transition is used as the objective function. 
Since the function is differentiable, 3D-NDT can directly optimize the function using gradient descending methods such as Gauss-Newton, which is more efficient than the ICP.
However, the high-resolution voxel-based representation can be computationally costly and somehow redundant especially in outdoor scenes. 
% Compared with ICP, it does not need to calculate the expensive nearest neighbor, and the probability density function can be calculated off-line between two frames.

%The advantage of NDT algorithm is that it does not need to consume a large amount of cost to calculate the nearest neighbor search matching point, and the probability density function can be calculated offline within two frames of point cloud acquisition time. 
%However, because too many grids need to be calculated in the outdoor scene, it is difficult to ensure efficiency and provide position estimates for an autonomous vehicle in real time.

%

Besides, the feature-based LO method was proposed for high-efficiency  \cite{zhang2014loam}. 
This method extracts planar and corner features for fast matching and achieves good results in various scenes when integrated with an IMU.
However, these features can be sparse in road dominated autonomous driving scenes.
And without IMU, LOAM tend to drift as shown in \Refsec{sec:exp}.
%its characteristic points are mainly angular and plane. This method has yielded good results in the urban road environment, but the effect is not common in the non-urban road environment，but it is not particularly desirable in non-urban road environments.

%

Similar to our method, the localization method for the outdoor environment based on a 2.5D representation was proposed in  \cite{wolcott2015fast}.
This method employs a multi-layer Gaussian mixture map (GMM) for describing the height variations in a 2D grid map.
Each grid cell in the grid map is modeled by a one-dimensional GMM of the height, and all grid maps are stitched in offline to form a prior localization map.
Such a representation is shown to be robust to environmental changes.
However, this method can only be used for online localization based on the prior map.
In addition, the EM algorithm is inefficient.

\begin{figure}
\centering
\includegraphics[height = 1.6in]{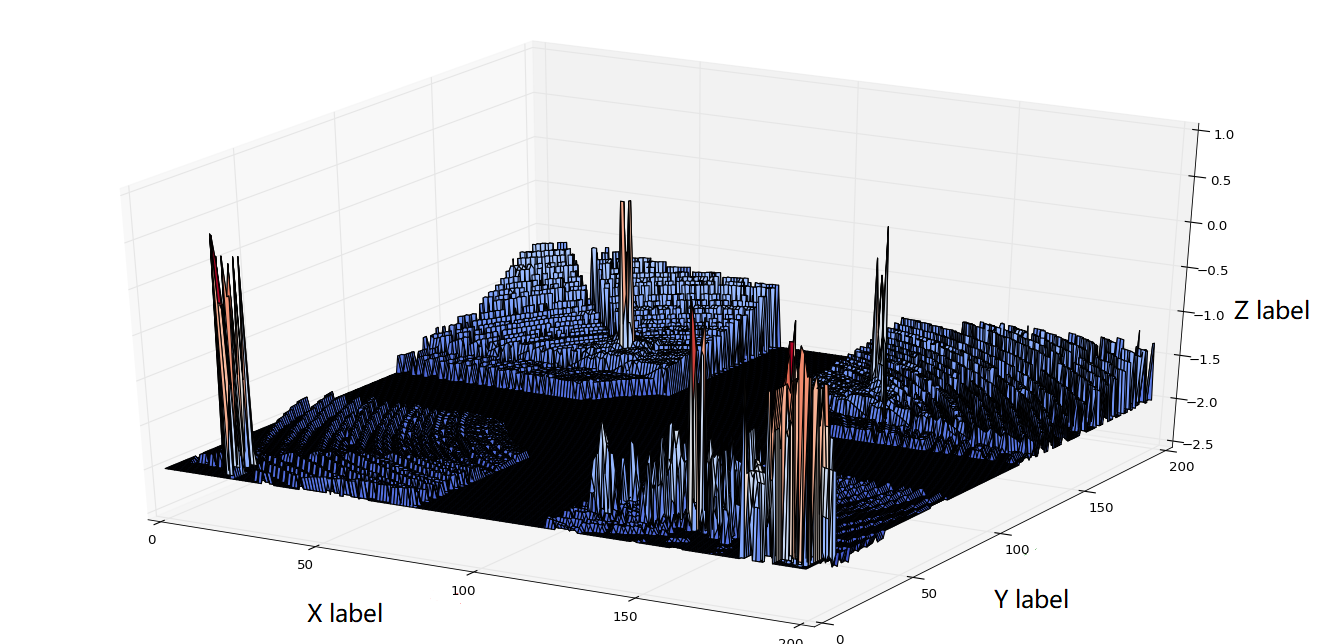}
\caption{
2.5D grid map example
}\label{fig:25D}
\end{figure}
\section{METHOD}
\subsection{2.5D grid map}  
We simplify the GMM-based 2.5D grid map proposed in  \cite{wolcott2015fast} by leaving only the expectation of heights in each grid.
Such a representation is much more efficient than the GMM-based map and can also indicate the height variation, which we found to be effective in the following matching process.
% Because 2.5D grid map not only contains the traditional 2D grid's advantages, with adding the height information but also take into account the advantages of 3D maps. It is very suitable for the map of the LO. However, due to the Gaussian mixture models 2.5D grid maps proposed in paper \cite{wolcott2015fast} are less efficient at registration. We simplify it, leaving only one average height in one grid instead of the Gaussian mixture model.
\Reffig{fig:25D} is an example of the 2.5D grid map.
The point cloud data should first be rasterized in a 2D grid map. 
Let the spherical coordinates of the i-th LiDAR point $p_i$ as $(\gamma_i,\phi_i,\theta_i)$, where $\gamma$ is the depth of the point, $\phi$ is the horizontal angle, and $\theta$ is the vertical angle. The 3D Cartesian coordinates of $p_i(x_i,y_i,z_i)$ is:
\begin{equation}
x_i=\gamma_i\cdot cos\theta_i\cdot cos\phi_i
\end{equation}
\begin{equation}y_i=\gamma_i\cdot cos\theta_i\cdot sin\phi_i\end{equation}
\begin{equation}z_i=\gamma_i\cdot sin\theta_i\end{equation}

We project the point cloud onto the 2D grid map. 
Here we use $C$ to represent the projection process of the point cloud, $(x_i^{'},y_i^{'})$ is the coordinate of $p_i$ in the grid map:
\begin{equation}Cp_i=p_{i}^{'}(x_i,y_i )\end{equation}
in which:
\begin{equation}x_i^{'}=floor(x_i/f_x+c_x)\end{equation}
\begin{equation}y_i^{'}=floor(y_i/f_y+c_y)\end{equation}
where $f_x$, $f_y$ are the grid map resolution in x and y axises, $c_x$, $c_y$ are the center of the grid map, which are half of the numbers of rows and columns.
The grid map is shown in \Reffig{fig:gridmap}(a).
For each grid, we keep the height $z$ of all points in the grid. 
Then the mean $\mu$ is used to represent the height expectation for each grid:
% Assume that the set of height z of all LiDAR points in the grid (i,j) is $g_{i, j}$, the mean of which is $\mu_{i, j}$. According to the mean definition
\begin{equation}\mu(p_i)=\frac{1}{n}\sum_{k=1}^{n}{z_k},\qquad n\ge3\end{equation}
where n is the number of points in $p_{i}^{'}$. Thus a 2.5D grid map projected by the point cloud is obtained. 
This map retains the advantages of easy storage and building as a 2D grid map, and also represent the height information which is crucial for registration in the outdoor environment.
It is for sure that a more precise model of height distribution can be further adopted in this 2.5D grid map, such as using normal distribution or GMM.
  
 % but also retains the point cloud height information. It is a kind of map representation between 2D and 3D, so is termed the 2.5D grid map.

\subsection{registration}   
\begin{figure}
\centering
\includegraphics[height = 1.6in]{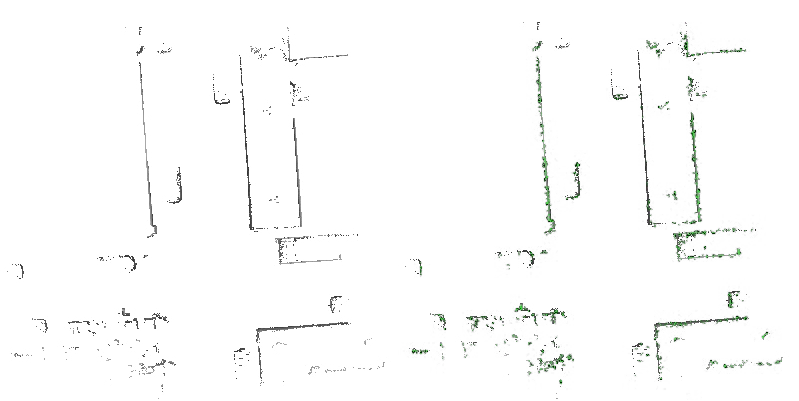}
\caption{
Grid maps show: (a) LiDAR points projected into the grid. (b) Grids with high gradient colored in green
}\label{fig:gridmap}
\end{figure}

Since the 2.5D grid map is analogous to "texture-less" gray-scale image as shown in \Reffig{fig:gridmap}, we adopt the direct method originated from VO to register two 2.5D grid maps  \cite{Gao2017SLAM}.
% In the VO, if the descriptor is not calculated during the matching process, all the pixels or feature point pixels are directly used to calculate the camera motion, which is known as the direct method\. 
In the direct method, 
% it is necessary to know the pixel positions of the spatial points projected into the camera coordinate. 
The optimization is no longer the reprojection error used in indirect feature-based matching, but the photometric error, which is the difference between the gray values of the same positions in two frames.
% : If using sparse feature points to calculate the error, known as the sparse direct method; if using a part of pixels, known as the semi-dense direct method; If using all pixels, known as the Dense Direct Method.

%

The direct method is based on the illumination invariant assumption: the gray value of the same spatial point is constant in sequent images. 
For cameras, this assumption does not always hold for all scenarios. 
However, in our case, this assumption is interpreted into the height invariant assumption, which is true in most of the time since the average height of a grid in the 2.5D grid map does not change, no matter how the autonomous vehicle moves. 
The compensation of pitch angle is crucial in practice to settle height disturbance caused by vehicle maneuvers.

Since most of the grids represent the ground with the height gradient be 0, only the grids with the height gradient greater than a certain threshold are used to calculate the height difference error (HDE). 
Therefore, the semi-dense direct method is suitable for 2.5D grid map registration for its efficiency. 
\Reffig{fig:gridmap}(b) shows the grids with high gradient extracted, which are colored in green.

%
% According to the number of pixels used, the direct method is further divided into sparse, dense and semi-dense.
We use the Gauss-Newton method to solve the transition matrix T of two 2.5D grid maps, the HDE between two grid maps is minimized. 
The objective function J is
\begin{equation}{\arg\min}_{R,t} J=\sum_{i=1}^{N}(\mu_1 (Cp_i )-\mu_2 (C(Rp_i+t)))^2 \end{equation}
where $\mu_1(p)$ and $\mu_2(p)$ represent the average height in grid $p$, $R$ is the rotation matrix and $t$ is the translation vector. Because transition matrix can't be added directly, we can't access to the partial derivative of $J$ with respect to $T$. 
As a result, Lie algebra $\xi$ is used to express the pose: 
\begin{equation}T=\begin{bmatrix}
R & t\\ 
0^T & 1
\end{bmatrix}
=exp(\xi^\land)\end{equation}
Then the objective function becomes:
\begin{equation}{\arg\min}_{\xi} J=\sum_{i=1}^{N}(\mu_1 (Cp_i )-\mu_2 (Cexp(\xi^\land)p_i))^2 \end{equation}
And the HDE $e_i$ is modeled as:
\begin{equation}e_i =\mu_1 (Cp_i )-\mu_2 (Cexp(\xi^\land)p_i)\end{equation}
To employ the Guass-Newton method, we need to calculate the Jacobian matrix $J_i$, which is the 
derivation of the independent variable $\xi$ by HDE $e_i$:
\begin{equation}J_i=\frac{\delta e_i}{\delta\xi}\end{equation}
suppose:
\begin{equation}q=exp(\xi^\land)p\end{equation}
\begin{equation}g=Cq\end{equation}
Because derivation of the $\xi$ by $e_i$ cannot be calculated directly, according to the chain rule, the $J_i$ can be divided into three parts:
\begin{equation}J_i=-\frac{\delta \mu_2}{\delta g}\frac{\delta g}{\delta q}\frac{\delta q}{\delta\xi}\end{equation}
From left to right: $\frac{\delta \mu_2}{\delta g}$ is the partial derivative of grid, which is the grid gradient, $\frac{\delta g}{\delta q}$ is the partial derivative of points in 3D space by grid, and $\frac{\delta q}{\delta\xi}$ is the partial derivative of Lie algebra by points in 3D space.

The grid gradient can be calculated as:
\begin{equation}\label{eq:16}
[(\mu_2 (u+1,v)-\mu_2 (u,v), \mu_2(u,v+1)-\mu_2(u,v))]
\end{equation}
We use bilinear interpolation to calculate floating-point coordinates of [u, v].

The left two partial derivatives can be calculated by:
$$
\frac{\delta g}{\delta q}\frac{\delta q}{\delta\xi}=
\begin{bmatrix}
f_x & 0 & -{f_x}x\\ 
0 & f_y & -{f_x}y
\end{bmatrix}
\begin{bmatrix}
1 & 0 & 0 & 0 & -1 & y\\
0 & 1 & 0 & 1 & 0 & -x\\
0 & 0 & 1 & -y & x & 0
\end{bmatrix}
$$
\begin{equation}\label{eq:17}
=
\begin{bmatrix}
f_x & 0 & {f_x}x & {f_x}xy & -f_x-{f_x}x^2 & {f_x}y\\
0 & f_y & -{f_y}y & f_y-{f_y}y & -{f_y}xy & -{f_y}y\\
\end{bmatrix}
\end{equation}
According to the equation (\ref{eq:16}) and equation (\ref{eq:17}), Jacobian matrix $J_i$ can be calculated, and Gauss-Newton method is used to optimize the objective function. 
The Gauss-Newton method is finally solved as:
\begin{equation}(\sum_{i=1}^{N}J_i^T J_i)\delta \xi^*=-\sum_{i=1}^{N}J_i^T e_i\end{equation}

% We can quickly iterate the transition matrix $T$ between these two frames.
We can then obtain the autonomous vehicle's position in every frame through continuous registration between frames which is termed as the Direct LiDAR Odometry (DLO).

\section{EXPERIMENTS}
\label{sec:exp}
\begin{figure}
\centering
\includegraphics[height = 2in]{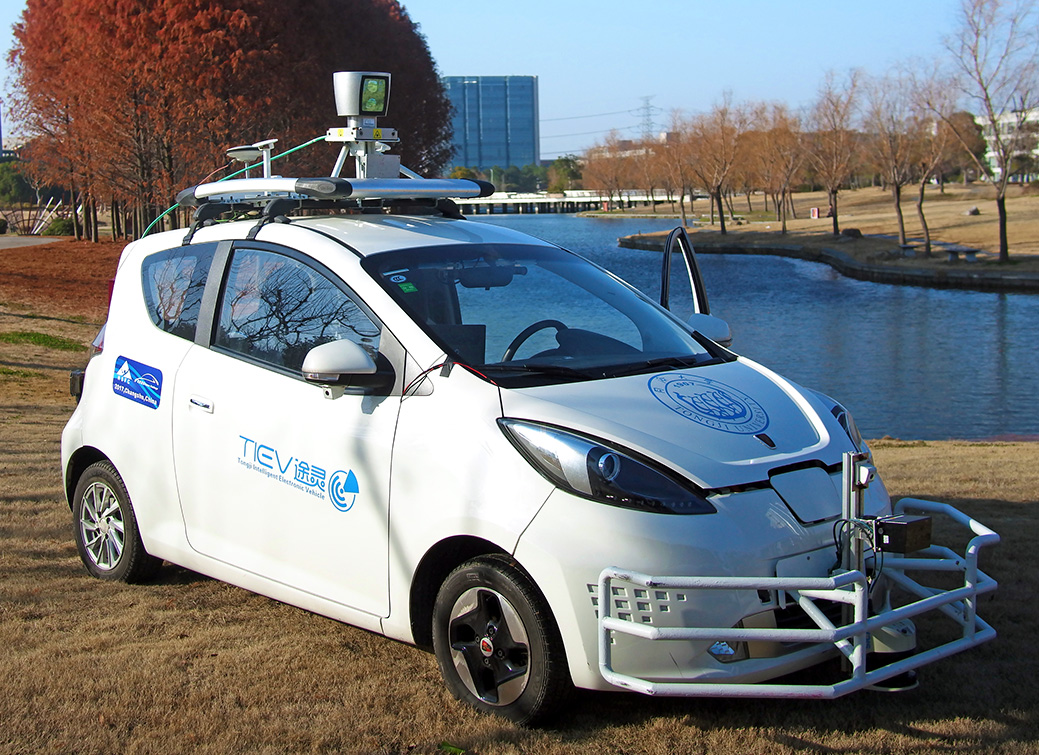}
\caption{
The TiEV autonomous vehicle platform
}\label{fig:Tiev}
\end{figure}

We used the KITTI dataset and datasets captured by the TiEV\footnote{cs1.tongji.edu.cn/tiev} autonomous driving platform, shown in (\Reffig{fig:Tiev}) to verify the proposed algorithm. 
TiEV equips sensors including HDL-64, VPL-16, IBEO, SICK, vision, RTKGPS + IMU. 
In this experiment, we used HDL-64 LiDAR only to collect data at Jiading campus of Tongji University to test our method.
In these experiments, the size of 2.5D grid map is 400 by 400, and each grid is 10cm by 10cm.  
The parameters in the algorithm were set as:  
$$f_x = 0.1$$
$$f_y = 0.1$$
$$c_x = 200$$
$$c_y = 200$$
The computing platform we use is a Core i5 4200U processor and 8GB memory. The operating system is Ubuntu 14.04.
\subsection{KITTI dataset}
KITTI \cite{Geiger2012CVPR} is a popular autonomous vehicle dataset in which the odometry data contains raw data from the camera, LiDAR, and poses of each frame. 
We used the point cloud data to test the proposed method and compare it with the ground truth. 
We also used open implementations of 3D-NDT\footnote{http://pointclouds.org/} and LOAM\footnote{https://github.com/laboshinl/loam\_velodyne} for comparison. 
Because DLO does not use the information of the IMU, we ran the LOAM program without the IMU data. 
\begin{figure}[ht]
\centering
\includegraphics[height = 2.2in]{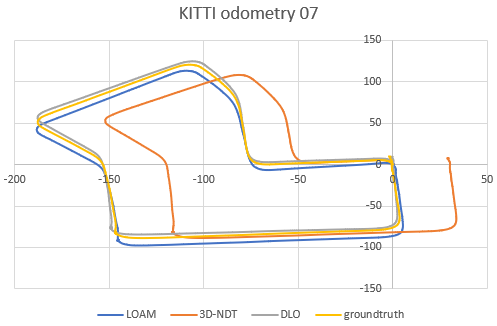}
\caption{Comparison between trajectories of DLO, 3D-NDT, LOAM (without IMU) and the ground truth using KITTI odometry dataset sequence 07
}\label{fig:kitti7}
\end{figure}

\begin{figure}[ht]
\centering
\includegraphics[height = 1.9 in]{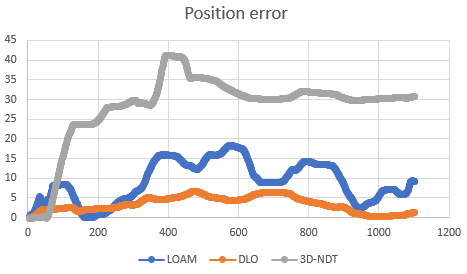}
\caption{
Comparison between position errors of DLO, 3D-NDT, LOAM (without IMU) using KITTI odometry dataset sequence 07
}\label{fig:position_error}
\end{figure}

\begin{figure}[ht]
\centering
\includegraphics[height = 2 in]{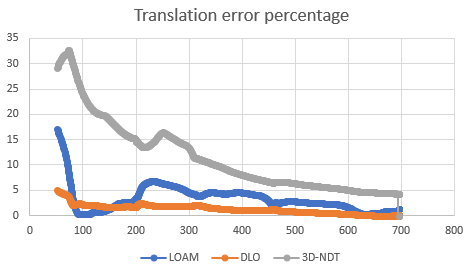}
\caption{
Comparison between Translation error percentages of DLO, 3D-NDT, LOAM (without IMU) using KITTI odometry dataset sequence 07
}\label{fig:Translation_error}
\end{figure}

\Reffig{fig:kitti7} shows the trajectories of the sequence 07 of the KITTI odometry dataset produced by DLO, 3D-NDT, LOAM and the ground truth.
This sequence is composed of 1100 frames and the full length is over 700m. 
% It is a closed curve on a plane，so it is very suitable for the experiments.
The result of DLO best conformed to the ground truth, with the absolute position error constantly below 3 meters as shown by \Reffig{fig:position_error}, which depicts deviations (error in meters) from the trajectory point to the ground truth point at each frame. 
% In \Reffig{fig:kitti7}, the X-axis are frames from 0 to 1100 and the Y-axis is the position error.
The error of 3D-NDT was considerably large, and LOAM is a little away from the groundtruth.
%For LOAM, big drifts were observed at corners, causing the overall positions shifted. 
However, LOAM still performed worse than DLO in this case. 
% \Reffig{fig:position_error} show the absolute position error at each frame
% The error, in the end, is over 25m. 
The corresponding error percentage is shown in \Reffig{fig:Translation_error}, which depicts the ratio between position error and the trajectory length, starting from 50 meters. 

\Reftab{tab:time} shows the running time of three methods on the dataset. 
It shows that DLO was much more efficient than 3D-NDT and LOAM as expected. 
3D-NDT used all point cloud information for scan matching and optimized the object function up to convergence, which resulted in huge computational time. 
LOAM is known to be able to yield good results based on the VLP-16 lidar. 
However, when dealing with the denser HDL-64 data, it is necessary to decrease the number of edge and planar points and to down-sample the point cloud.
The frame rate after careful tuning is at most 1hz (including both matching and mapping, the matching was conducted at 3hz), which is not comparable to DLO.
% It can be concluded that DLO is a much better than NDT and LOAM without IMU, both in accuracy and efficiency.
\begin{table} [ht]
\caption{
Running times of DLO, 3D-NDT, and LOAM
}
\centering
\begin{tabular}{|c|c|c|c|}% 通过添加 | 来表示是否需要绘制竖线
\hline  % 在表格最上方绘制横线
&DLO&3D-NDT&LOAM\\
\hline  %在第一行和第二行之间绘制横线
total time &44s&43153s&1025s\\
\hline % 在表格最下方绘制横线
time/frame &0.04s&39.23s&0.93s\\
\hline % 在表格最下方绘制横线
\end{tabular}
\label{tab:time}
\end{table}

\begin{figure}
\centering
\includegraphics[height = 2 in]{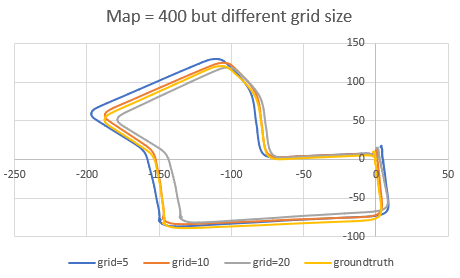}
\caption{
The different size of the grid based on DLO
}\label{fig:gridsize}
\end{figure}

We further compared results generated based on different sizes of the grid in \Reffig{fig:gridsize}. 
We concluded that the size of 10cm by 10cm is a good choice considering both the accuracy and the time-consumption.

\subsection{The dataset of Tongji university} 
\begin{figure}[h]
\centering
\includegraphics[height = 2 in]{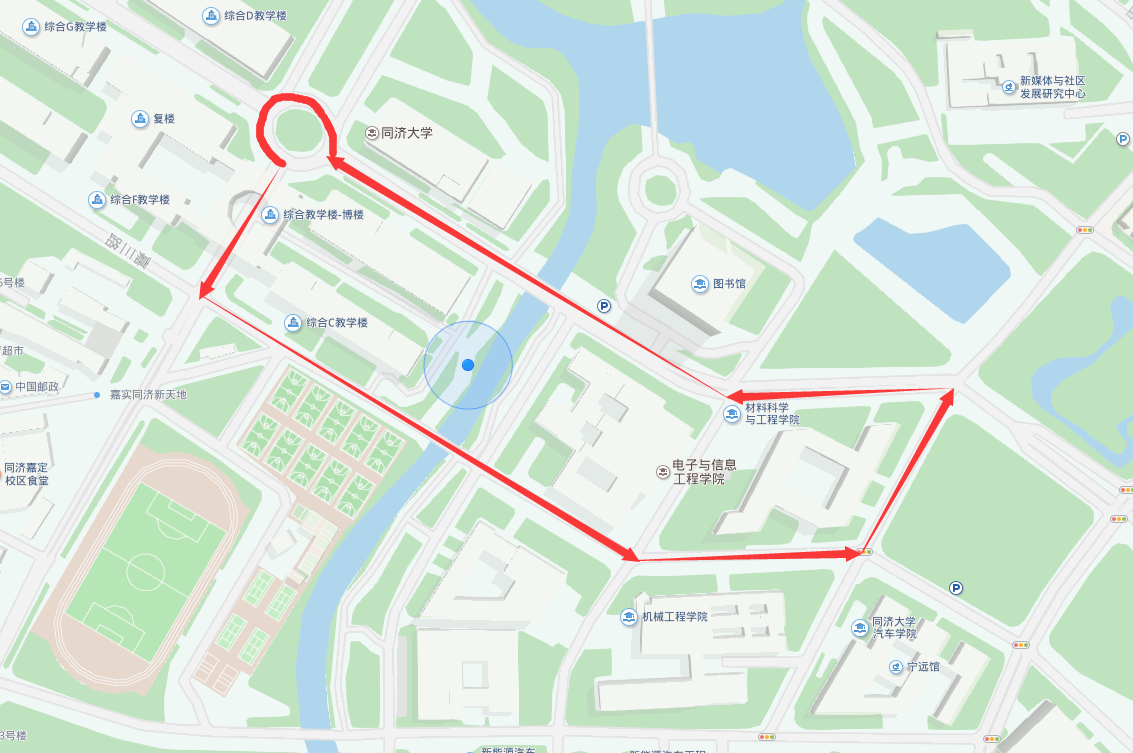}
\caption{
The experimental route in the Jiading dataset
}\label{fig:map}
\end{figure}

We also tested DLO using dataset captured by TiEV as shown in \Reffig{fig:map}.
This is a closed outdoor route with a total length of about 1.3km, including turns with vagarious angles of about 60$^{\circ}$, 90$^{\circ}$, 120$^{\circ}$, and a 270$^{\circ}$ turn in a roundabout.
This dataset contains more than 2000 frames.
During the experiment, we found the 3D-NDT took days to produce a result, so we compared DLO only with LOAM using this dataset. 

\begin{figure}[ht]
\centering
\includegraphics[height = 2 in]{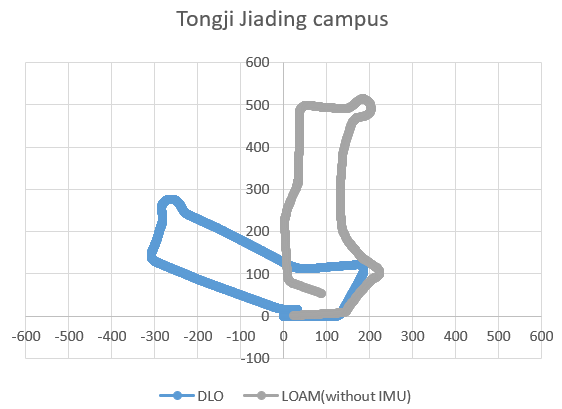}
\caption{
Comparison of the trajectories of DLO and LOAM (without IMU) using Jiading dataset
}\label{fig:jiading-2}
\end{figure}

In \Reffig{fig:jiading-2}, LOAM resulted in an inaccurate trajectory either in straight paths or at corners. 
% In the corner part, the angle of registration is larger than the angle of the corner, resulting in the trajectory completely deviate from the desired shape. 
The trajectory deviates from the starting point at around 200 meters at the end. 
% The distance between the starting point and the ending point is around 239m.
The proposed DLO performed much better. However, it is also not completely closed. 
The distance between the starting point and the ending point is about 32 m, far less than the LOAM result without IMU.
We found the error occurred mostly at corners which can be improved if IMU data is integrated.
Moreover, the extra cost for estimating a 6-DoF pose in 3D-NDT and LOAM is restricted since all the used datasets comprise relative flat area.
Therefore, it is appropriate to estimate the 3 DoF pose instead of the 6 DoF pose.

% , the truth is that the z value keeps constant when our car is driving on a plane, in other words, it is fair that the computational time and the final result of DLO which is based on the 6 DoF pose can be compared with NDT and LOAM.
% The error in the straight part is very small, the distance estimation is very accurate. However, there is a slight error in the corner turns, resulting in the two lines that should be parallel not completely parallel. 
% So while LOAM works well with IMU, it relies too much on the initial value provided by IMU. In the absence of IMU information, DLO can achieve better results than LOAM. Compared to LOAM, DLO only calculates motion in a two-dimensional plane and relies less on initial values. DLO also showed no disadvantage regarding runtime, and both methods can run in real time.

\section{CONCLUSIONS}

In this paper, a DLO method based on the 2.5D grid map is proposed.
This method is highly efficient and suitable for the outdoor environment. 
The point cloud of a frame is firstly projected onto a 2.5D grid map and then registered using the semi-dense direct method. 
Experiments with the KITTI dataset and the dataset we captured demonstrated that DLO outperforms 3D-NDT, both in accuracy and efficiency. 
Compared with the LOAM, which is the state-of-the-art in the KITTI benchmark list, DLO does not reply on extracted features from the point cloud. Therefore, the registration accuracy will be significantly higher in structureless scenes.
The shortcoming of this method is the assumption of a flat world, which however makes sense for autonomous driving tasks.
Another possible improvement is to adopt the height distribution instead of the height expectation which will be explored in our future work.

\bibliographystyle{IEEEtran}
\bibliography{IEEEabrv,ref}

\begin{thebibliography}{10}
\providecommand{\url}[1]{#1}
\csname url@rmstyle\endcsname
\providecommand{\newblock}{\relax}
\providecommand{\bibinfo}[2]{#2}
\providecommand\BIBentrySTDinterwordspacing{\spaceskip=0pt\relax}
\providecommand\BIBentryALTinterwordstretchfactor{4}
\providecommand\BIBentryALTinterwordspacing{\spaceskip=\fontdimen2\font plus
\BIBentryALTinterwordstretchfactor\fontdimen3\font minus
  \fontdimen4\font\relax}
\providecommand\BIBforeignlanguage[2]{{%
\expandafter\ifx\csname l@#1\endcsname\relax
\typeout{** WARNING: IEEEtran.bst: No hyphenation pattern has been}%
\typeout{** loaded for the language `#1'. Using the pattern for}%
\typeout{** the default language instead.}%
\else
\language=\csname l@#1\endcsname
\fi
#2}}

\bibitem{engel2017direct}
J.~Engel, V.~Koltun, and D.~Cremers, ``Direct sparse odometry,'' \emph{IEEE
  Transactions on Pattern Analysis and Machine Intelligence}, pp. 1--1, 2017.

\bibitem{mur2015orb}
R.~Mur-Artal, J.~M.~M. Montiel, and J.~D. Tardos, ``Orb-slam: a versatile and
  accurate monocular slam system,'' \emph{IEEE Transactions on Robotics},
  vol.~31, no.~5, pp. 1147--1163, 2015.

\bibitem{engel2014lsd}
J.~Engel, T.~Sch{\"o}ps, and D.~Cremers, ``Lsd-slam: Large-scale direct
  monocular slam,'' in \emph{European Conference on Computer Vision}.\hskip 1em
  plus 0.5em minus 0.4em\relax Springer, 2014, pp. 834--849.

\bibitem{forster2014svo}
C.~Forster, M.~Pizzoli, and D.~Scaramuzza, ``Svo: Fast semi-direct monocular
  visual odometry,'' in \emph{Robotics and Automation (ICRA), 2014 IEEE
  International Conference on}.\hskip 1em plus 0.5em minus 0.4em\relax IEEE,
  2014, pp. 15--22.

\bibitem{Lawler1966Branch}
E.~L. Lawler and D.~E. Wood, ``Branch-and-bound methods: A survey,''
  \emph{Operations Research}, vol.~14, no.~4, pp. 699--719, 1966.

\bibitem{grisetti2007improved}
G.~Grisetti, C.~Stachniss, and W.~Burgard, ``Improved techniques for grid
  mapping with rao-blackwellized particle filters,'' \emph{IEEE transactions on
  Robotics}, vol.~23, no.~1, pp. 34--46, 2007.

\bibitem{KohlbrecherMeyerStrykKlingaufFlexibleSlamSystem2011}
S.~Kohlbrecher, J.~Meyer, O.~von Stryk, and U.~Klingauf, ``A flexible and
  scalable slam system with full 3d motion estimation,'' in \emph{Proc. IEEE
  International Symposium on Safety, Security and Rescue Robotics
  (SSRR)}.\hskip 1em plus 0.5em minus 0.4em\relax IEEE, November 2011.

\bibitem{hess2016real}
W.~Hess, D.~Kohler, H.~Rapp, and D.~Andor, ``Real-time loop closure in 2d lidar
  slam,'' in \emph{Robotics and Automation (ICRA), 2016 IEEE International
  Conference on}.\hskip 1em plus 0.5em minus 0.4em\relax IEEE, 2016, pp.
  1271--1278.

\bibitem{besl1992method}
P.~J. Besl, N.~D. McKay, \emph{et~al.}, ``A method for registration of 3-d
  shapes,'' \emph{IEEE Transactions on pattern analysis and machine
  intelligence}, vol.~14, no.~2, pp. 239--256, 1992.

\bibitem{magnusson2009three}
M.~Magnusson, ``The three-dimensional normal-distributions transform: an
  efficient representation for registration, surface analysis, and loop
  detection,'' Ph.D. dissertation, {\"O}rebro universitet, 2009.

\bibitem{zhang2014loam}
J.~Zhang and S.~Singh, ``Loam: Lidar odometry and mapping in real-time.'' in
  \emph{Robotics: Science and Systems}, vol.~2, 2014.

\bibitem{wolcott2015fast}
R.~W. Wolcott and R.~M. Eustice, ``Fast lidar localization using
  multiresolution gaussian mixture maps,'' in \emph{Robotics and Automation
  (ICRA), 2015 IEEE International Conference on}.\hskip 1em plus 0.5em minus
  0.4em\relax IEEE, 2015, pp. 2814--2821.

\bibitem{Gao2017SLAM}
X.~Gao, T.~Zhang, Y.~Liu, and Q.~Yan, \emph{14 Lectures on Visual SLAM: From
  Theory to Practice}.\hskip 1em plus 0.5em minus 0.4em\relax Publishing House
  of Electronics Industry, 2017.

\bibitem{Geiger2012CVPR}
A.~Geiger, P.~Lenz, and R.~Urtasun, ``Are we ready for autonomous driving? the
  kitti vision benchmark suite,'' in \emph{Conference on Computer Vision and
  Pattern Recognition (CVPR)}, 2012.

\end{thebibliography}

\end{document}